\documentclass[11pt]{article}
\usepackage{fullpage,graphicx,psfrag,amsmath,amsfonts,verbatim}
\usepackage{xcolor}
\usepackage{amsthm}
\usepackage[small,bf]{caption}
\usepackage{authblk}

\usepackage{amsmath}
\usepackage{amssymb}
\usepackage{mathtools}
\usepackage{amsthm}
\usepackage{amsfonts}
\usepackage{algorithm}
\usepackage{algorithmic}

\usepackage{microtype}
\usepackage{graphicx}
\usepackage{subfigure}
\usepackage{booktabs} 
\usepackage{hyperref}
\allowdisplaybreaks

\title{A Survey of Automatic Prompt Engineering: An Optimization Perspective}
\author[1]{Wenwu Li}
\author[4,3]{Xiangfeng Wang}
\author[1]{Wenhao Li\thanks{\texttt{whli@tongji.edu.cn}}}
\author[2,1]{Bo Jin\thanks{\texttt{bjin@tongji.edu.cn}}}

\affil[1]{School of Computer Science and Technology, Tongji University}
\affil[2]{Shanghai Research Institute for Intelligent Autonomous Systems, Tongji University}
\affil[3]{School of Computer Science and Technology, East China Normal University}
\affil[4]{Key Laboratory of Mathematics and Engineering Applications, MoE, East China Normal University}

\date{}
\begin{document}
\maketitle

\begin{abstract}
The rise of foundation models has shifted focus from resource-intensive fine-tuning to prompt engineering, a paradigm that steers model behavior through input design rather than weight updates.
While manual prompt engineering faces limitations in scalability, adaptability, and cross-modal alignment, automated methods, spanning foundation model (FM) based optimization, evolutionary methods, gradient-based optimization, and reinforcement learning, offer promising solutions. 
Existing surveys, however, remain fragmented across modalities and methodologies. 
This paper presents the first comprehensive survey on automated prompt engineering through a unified optimization-theoretic lens. 
We formalize prompt optimization as a maximization problem over discrete, continuous, and hybrid prompt spaces, systematically organizing methods by their optimization variables (instructions, soft prompts, exemplars), task-specific objectives, and computational frameworks. 
By bridging theoretical formulation with practical implementations across text, vision, and multimodal domains, this survey establishes a foundational framework for both researchers and practitioners, while highlighting underexplored frontiers in constrained optimization and agent-oriented prompt design.
\end{abstract}

\section{Introduction}\label{sec:intro}


The transformative impact of pre-trained foundation models (FMs, e.g., large language models, LLMs or vision language models, VLMs) has revolutionized natural language processing and visual understanding, enabling unprecedented capabilities in complex cognitive tasks ranging from mathematical reasoning to multi-agent collaboration systems~\cite{xi2025rise}. 
As model scales escalate into the trillions of parameters, conventional fine-tuning approaches face prohibitive computational barriers. 
This resource intensiveness fundamentally restricts FM deployment in real-world applications, particularly for edge devices and time-sensitive scenarios like autonomous vehicle decision-making or real-time medical diagnosis.


Several efficiency-focused approaches have emerged including parameter-efficient fine-tuning, model distillation, sparse training, and dynamic architecture methods~\cite{wan2024efficient}. 
While these reduce computational demands to varying degrees, they maintain dependency on parameter updates requiring substantial training data and backpropagation mechanics. 
This proves particularly limiting in scenarios demanding rapid adaptation - such as financial fraud detection systems needing hourly model updates, or medical applications where patient data privacy prohibits retraining. 
Prompt engineering circumvents these constraints through a paradigm shift: rather than modifying neural weights, it reprograms FM behavior via strategic input design~\cite{sahoo2024systematic,amatriain2024prompt,vatsal2024survey}.


Manual prompt engineering has demonstrated remarkable generalizability through techniques like Chain-of-Thought (CoT) and few-shot exemplar selection~\cite{akinwande2024understanding}.
However, its practical adoption faces fundamental limitations: 
1) \textit{expert dependency} requiring laborious trial-and-error~\cite{sahoo2024systematic};
2) \textit{input format sensitivity} where minor syntactic variations (e.g., punctuation changes or instruction phrasing) yield performance fluctuations~\cite{sclar2024quantifying}, and
3) \textit{static design} unable to adapt to evolving inputs like shifting social media discourse patterns~\cite{zhou2022large}. 
These limitations intensify in multimodal systems where manual prompting must resolve cross-modal alignment challenges - for instance, ensuring visual grounding accuracy in vision language models (VLMs) requires precise coordination between image region descriptors and textual queries that humans often misalign~\cite{gu2023systematic}.

Automated prompt optimization addresses these limitations through systematic exploration of combinatorial prompt spaces using evolutionary strategies that mutate token sequences through genetic operations, reinforcement learning (RL) that treat prompts as differentiable policies, and meta-learning approaches that adapt prompts through gradient-based hyperparameter optimization. 
Crucially, these methods demonstrate emergent capabilities surpassing human design, such as automatically discovering prompts that balance multiple objectives~\cite{menchaca2025mopo} or adaptively reconfigure based on real-time feedback in robotics control systems~\cite{ma2024eureka}.


Existing surveys remain fragmented across methodological and modal boundaries. 
While foundational works establish prompt-based learning theory \cite{liu2023pre} and multimodal applications \cite{gu2023systematic}, specialized reviews focus on compression techniques \cite{li2024prompt} or manual design patterns \cite{sahoo2024systematic,amatriain2024prompt,vatsal2024survey}. 
Though \cite{chang2024efficient} addresses efficiency aspects, no comprehensive treatment exists for automated prompt engineering across modalities.


\begin{figure*}[htb!]
    \centering
    \includegraphics[width=\linewidth]{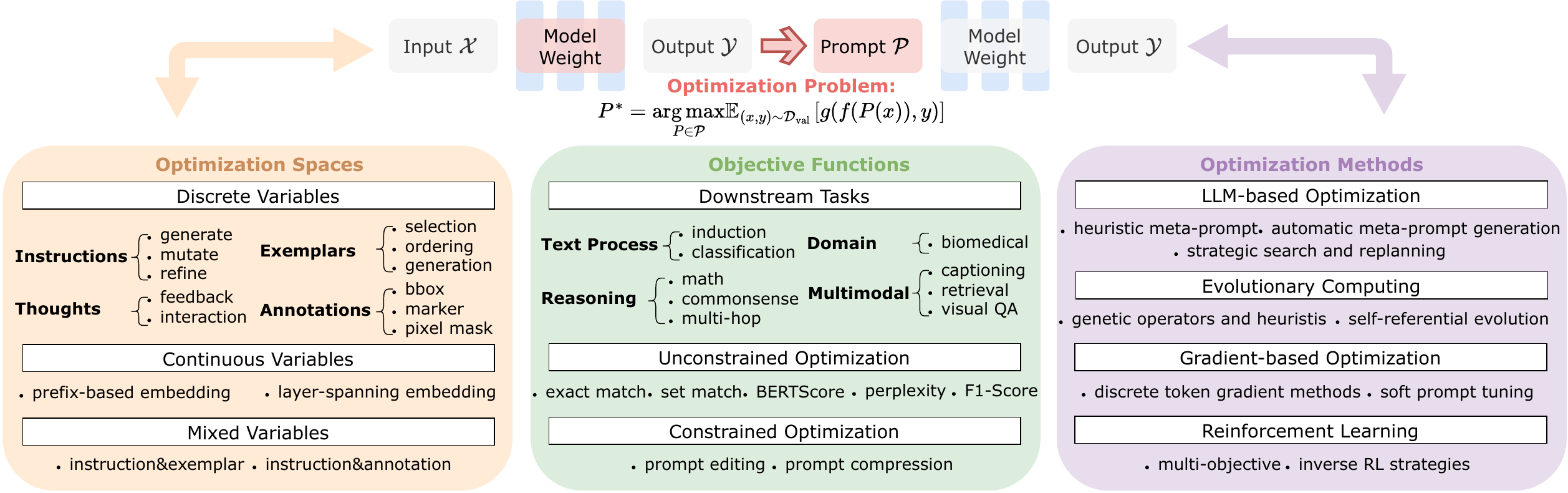}
    \caption{A unified optimization framework for the automated prompt engineering.}
    \label{fig:overview}
\end{figure*}

Our work establishes the first unified optimization theoretic framework (Figure~\ref{fig:overview}) for automated prompt engineering across modalities. 
We formalize the problem as maximizing expected performance metrics over discrete, continuous, and hybrid prompt spaces (Section~\ref{sec:formulation}), where different variable types (hard instructions, soft prompts, few-shot exemplars and mixed variables) correspond to specific optimization subproblems.
This survey systematically organizes existing methods through this lens: 
(1) Optimization Spaces (Section~\ref{sec:variable}) categorize prompt elements across text, vision, and multimodal domains; 
(2) Objective Functions (Section~\ref{sec: obj}) characterize various task categories with mathematical instantiations of performance metrics and constraints; 
(3) Optimization Methods (Section~\ref{sec: method}) classify techniques into representative computational paradigms (FM-based, evolutionary, gradient-based, and RL).
This unified view not only helps explain the effectiveness of existing methods but also establishes a rigorous foundation for developing more sophisticated automated prompt engineering algorithms, bridging the gap between theoretical understanding and practical implementation while identifying underexplored research frontiers.


\section{Related Work}\label{sec:related-work}

Prompt engineering has rapidly gained prominence as an essential method for enhancing the capabilities of FMs across diverse tasks. 
Several surveys have contributed to this domain by examining specific facets of prompt-based learning. 
For instance, \cite{liu2023pre} position prompt engineering within the broader context of natural language processing (NLP), emphasizing how textual prompts differ from traditional supervised training. 
Similarly, \cite{vatsal2024survey} and \cite{sahoo2024systematic} each provide comprehensive overviews of prompt methods (ranging from meticulously crafted natural language instructions to learned vectors), detailing their performance across various NLP benchmarks. 
Meanwhile, \cite{amatriain2024prompt} explore more advanced prompt-based mechanisms such as CoT and Reflection and examine toolkits like LangChain and Semantic Kernel.

In addition to these foundational surveys, domain- and technique-specific works have emerged. 
\cite{li2024prompt} focus on prompt compression methods, exploring both hard and soft approaches and illustrating the ways in which compression can streamline model performance without sacrificing accuracy. 
\cite{gu2023systematic} expand these ideas to vision-language models, distinguishing between multimodal-to-text generation, image-text matching, and text-to-image generation. 
They investigate how prompting strategies differ in multimodal settings compared to purely textual ones. 
Thus, these surveys collectively delineate a growing research landscape that spans multiple model types, task formats, and optimization considerations.

Yet, while these works are individually valuable, they remain fragmented by methodological or modal boundaries. 
\cite{chang2024efficient} specifically address efficiency aspects, but no comprehensive resource unifies the theoretical underpinnings across discrete, continuous, and hybrid prompt spaces. 
Existing surveys either concentrate on foundational theories~\cite{liu2023pre,gu2023systematic}, explore a narrowly defined technique such as compression~\cite{li2024prompt}, or focus on manual design patterns~\cite{sahoo2024systematic,vatsal2024survey}.
This compartmentalization leaves open questions about how to systematically organize prompt components, objectives, and optimization strategies under a cohesive theoretical framework.

Our survey bridges this gap by introducing a unified, optimization-theoretic perspective on automated prompt engineering across modalities. 
Unlike prior works that center on specific tasks, prompt types, or efficiency improvements, we formulate prompt engineering as an overarching optimization problem that seeks to maximize task-specific performance metrics for discrete (hard and exampler), continuous (soft), and mixed prompts. 
Our taxonomy spans variables, objective functions, and optimization methods, clarifying best practices and underscoring promising directions for future research. 
In doing so, we provide researchers and practitioners with a rigorous foundation for advancing automated prompt engineering, synthesizing the diverse strands of existing studies into a single, cohesive framework.

\section{Optimization Problem Formulation}\label{sec:formulation}

This paper studies the prompt optimization problem for foundation models, including both LLMs and VLMs. 
Let $\mathcal{X}$ denote the input space and $\mathcal{Y}$ denote the output space. 
For LLMs, $\mathcal{X}$ represents text inputs, while for VLMs, $\mathcal{X} = \mathcal{X}_v \times \mathcal{X}_t$ represents image-text pairs, where $\mathcal{X}_v$ denotes the visual space and $\mathcal{X}_t$ denotes the text space. 
A prompt function $P: \mathcal{X} \rightarrow \mathcal{P}$ maps input queries to a conditioning pattern that elicits specific model behaviors. 
The prompt space $\mathcal{P}$ can be partitioned into three subspaces: 
the discrete prompt space $\mathcal{P}_d$, the continuous prompt space $\mathcal{P}_c$, and the hybrid prompt space $\mathcal{P}_h = \mathcal{P}_d \times \mathcal{P}_c$.

For $P \in \mathcal{P}_d$, we consider different canonical forms based on model type. 
In LLMs, the zero-shot form is expressed as $P(x)=\left[I,T; x\right]$, where $I,T \in \mathcal{V}^*$ denotes a learnable instruction and thought sequence from vocabulary space $\mathcal{V}$ respectively, and $\mathcal{V}^*$ represents the set of all possible sequences over $\mathcal{V}$. 
The few-shot form $P(x)=\left[I, T, e_1, \ldots, e_k; x\right]$, where $\{e_i \in \mathcal{E}\}_{i=1}^k$ are $k$ learnable exemplars from space $\mathcal{E}$. 
Each exemplar $e_i = (x_i^e, y_i^e)$ consists of an I/O pair where $x_i^e \in \mathcal{X}$ and $y_i^e \in \mathcal{Y}$.

In VLMs, besides inheriting all prompt forms from LLMs, the spatial annotation form takes a general expression $P(x)=\left[I, T, R_1, \ldots, R_m; x\right]$, where $\{R_i \in \mathcal{R}\}_{i=1}^m$ are spatial regions from a general region space $\mathcal{R}$. 
Each region $R_i = (A_i, l_i)$ consists of an area specification $A_i \in \mathcal{A}$ and a label $l_i \in \mathcal{L}$, where $\mathcal{A}$ is a general area specification space that can represent various forms including but not limited to: 
1) bounding boxes: $A_i \in [0,1]^4$ representing normalized coordinates; 
2) markers: $A_i \in [0,1]^3$ representing center coordinates and radius; 
3) pixel masks: $A_i \in \{0,1\}^{H \times W}$ representing binary masks; and 
4) other region specifications (e.g., polygons, curves).

For $P \in \mathcal{P}_c$, the prompt takes a unified form:
\begin{equation}
P(e_x)=\left[\theta_1, \ldots, \theta_m; e_x\right],
\end{equation}
where $e_x = \text{Embed}(x) \in \mathbb{R}^d$ is the embedding representation of input $x$ through an embedding function $\text{Embed}: \mathcal{X} \rightarrow \mathbb{R}^d$, and $\{\theta_i \in \mathbb{R}^d\}_{i=1}^m$ are $m$ learnable vectors in the $d$-dimensional embedding space.
For $P \in \mathcal{P}_h$, the prompt combines both discrete and continuous elements:
\begin{equation}
P(x, e_x)=\left[I, T, R_1, \ldots, R_k, \theta_1, \ldots, \theta_m; x\right],
\end{equation}
allowing joint optimization of discrete regions and continuous embeddings.

\begin{table}[ht]
\centering
\resizebox{.7\linewidth}{!}{%
\begin{tabular}{lll}
\hline
\textbf{Symbol} & \textbf{Space} & \textbf{Definition} \\
\hline
$x$ & $\mathcal{X} = \mathcal{X}_v \times \mathcal{X}_t$ & Input query (text or image-text pair) \\
$y$ & $\mathcal{Y}$ & Output response \\
$\mathcal{V}$ & - & Vocabulary of tokens \\
$I$ & $\mathcal{V}^*$ & Instruction sequence \\
$T$ & $\mathcal{V}^*$ & Thought sequence \\
$e_i$ & $\mathcal{E} = \mathcal{X} \times \mathcal{Y}$ & Text exemplar (input-output pair) \\
$\mathcal{R}$ & - & General region space \\
$A_i$ & $\mathcal{A}$ & Area specification (box/marker/mask/etc.) \\
$R_i$ & $\mathcal{R} = \mathcal{A} \times \mathcal{L}$ & Spatial region with label \\
$\theta_i$ & $\mathbb{R}^d$ & Learnable prompt vector \\
$e_x$ & $\mathbb{R}^d$ & Input embedding \\
$P$ & $\mathcal{P} = \mathcal{P}_d \cup \mathcal{P}_c$ & Prompt function \\
$\mathcal{P}_d$ & $\mathcal{P}_d^t \times \mathcal{P}_d^v$ & Discrete prompt space \\
$\mathcal{P}_c$ & - & Continuous prompt space \\
$\mathcal{P}_h$ & $\mathcal{P}_d \times \mathcal{P}_c$ & Hybrid prompt space \\
$f$ & $\mathcal{P} \times \mathcal{X} \rightarrow \mathcal{Y}$ & Foundation model \\
$g$ & $\mathcal{Y} \times \mathcal{Y} \rightarrow \mathbb{R}$ & Performance metric \\
$\mathcal{D}_{\text{val}}$ & $(\mathcal{X} \times \mathcal{Y})^{n_{\text{val}}}$ & Validation dataset of size $n_{\text{val}}$ \\
\hline
\end{tabular}%
}
\caption{Summary of Mathematical Notation}
\end{table}

Given a black-box foundation model $f: \mathcal{P} \times \mathcal{X} \rightarrow \mathcal{Y}$ and a validation set $\mathcal{D}_{\text{val}}=\{(x_i, y_i)\}_{i=1}^{n_{\text{val}}}$ of size $n_{\text{val}}$, the prompt optimization problem can be formulated as:
\begin{equation}\label{eq:main-formula}
\begin{aligned}
P^* = & \underset{P \in \mathcal{P}}{\arg\max} & & \mathbb{E}_{(x,y) \sim \mathcal{D}_{\text{val}}}[g(f(P(x)), y)] \\
& \text{subject to} & & P \in \mathcal{P}_d \text{ or } P \in \mathcal{P}_c \text{ or } P \in \mathcal{P}_h,
\end{aligned}
\end{equation}
where $g: \mathcal{Y} \times \mathcal{Y} \rightarrow \mathbb{R}$ denotes a performance metric measuring the quality of model predictions against ground truth.
This formulation leads to $3$ subclasses of problems:
$$
\max_{P \in \mathcal{P}_\star} \mathbb{E}_{(x,y)}\left[g(f(P(\triangle)), y)\right],
\begin{cases}
\star = d,\triangle = x,&\text{(DPO)} \\
\star = c,\triangle = e_x,&\text{(CPO)} \\
\star = h,\triangle = (x, e_x),&\text{(HPO)}
\end{cases}
$$
where DPO, CPO, and HPO denote discrete, continuous, and hybrid prompt optimization respectively. 
Solutions for all cases are discussed in subsequent sections.

\section{Optimization Spaces}\label{sec:variable}

\begin{figure}[htb!]
    \centering
    \includegraphics[width=.6\linewidth]{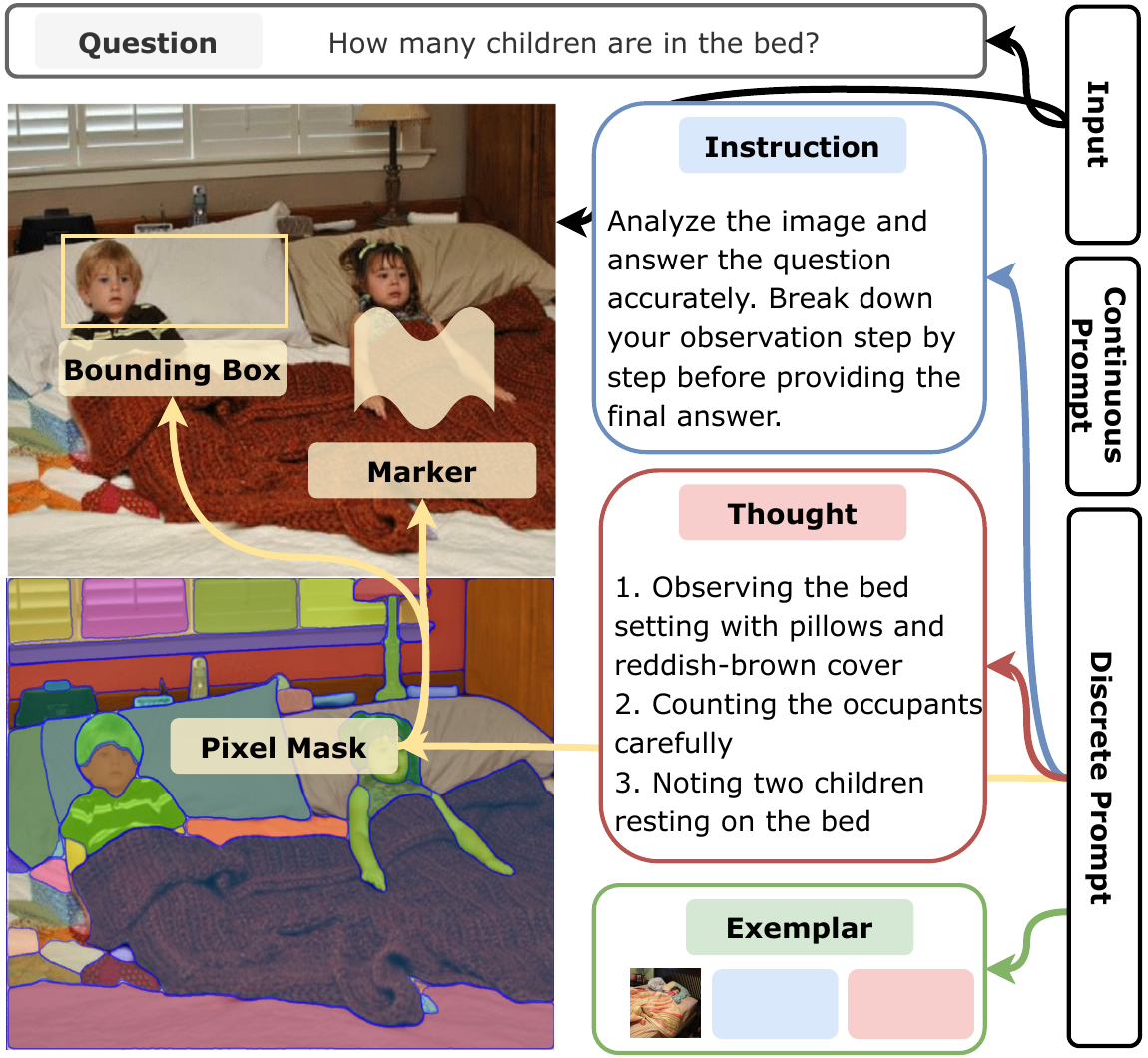}
    \caption{A prompt example for the visual question-answering task.}
    \label{fig:prompt-example}
\end{figure}

Building upon formulation in Section~\ref{sec:formulation}, we systematically analyze the $3$ fundamental variable types in prompt optimization: 
discrete ($\mathcal{P}_d$), continuous ($\mathcal{P}_c$), and hybrid combinations ($\mathcal{P}_h$) (Figure~\ref{fig:prompt-example}). 
This taxonomy enables principled analysis of automated prompt engineering across modalities.

\subsection{Discrete Variables}

Discrete prompt optimization operates in $\mathcal{P}_d$ space, manipulating human-interpretable elements through combinatorial search. 
We identify three principal variable subtypes:

\paragraph{Instructions}\label{sec:hard-instruction}



Instruction variables ($I \in \mathcal{V}^*$) specify task objectives through natural language directives (e.g., ``Translate to French''). 
Their optimization centers on generating concise yet precise directives to elicit strong model performance. 
Representative approaches \textit{generate}~\cite{hsieh2023automatic,yang2023largelm,wang2023promptagent}, \textit{mutate}~\cite{yang2023largelm}\cite{guo2023connecting} or \textit{refine}~\cite{shin2020autoprompt,pryzant2023automatic,guo2023connecting,kong2024prewrite,kwon2024stableprompt,deng2022rlprompt,hu2024localized} instructions via gradient-inspired edits, GA, or RL to systematically improve task accuracy.
This classification can be further extended to multimodal prompts, with representative approaches including  \emph{generate}~ \cite{kim2023multiprompter}, \emph{mutate}~ \cite{hao2024optimizing,ogezi2024optimizing} or 
 \emph{refine}~ \cite{manas2024improving,yeh2024tipo,mrini2024fast,wang2024promptcharm,rosenman2023neuroprompts,mo2024dynamic}, which play a crucial role in optimizing accuracy and ensuring consistency in generated image quality.

\paragraph{Thoughts}

Thought variables ($T \in \mathcal{V}^*$) implement chain-of-reasoning through intermediate reasoning steps.
By decomposing large problems into smaller subproblems, CoT guidance can substantially boost solution correctness when automatically generated or refined.
These approaches enhance the precision and flexibility of CoT reasoning, enabling models to better handle complex tasks.
Representative methods include \emph{feedkback}~\cite{press2023measuring,chen2024prompt,chen2024reprompt}, which uses self or external feedback to refine reasoning, and \emph{interaction}~\cite{yaoreact,madaan2024self}, which leverages external tools or resources to verify and adjust reasoning dynamically.

\paragraph{Few-shot Examples} \label{sec:example}

These short input-output pairs (${e_i} \subset \mathcal{E}$) demonstrate task-relevant behaviors, assisting models in inferring patterns with minimal supervision. 
Exemplar optimization techniques may address exemplar \emph{selection}~\cite{liu2022makes,margatina2023active,zhang2022active} or \emph{ordering} \cite{lu2022fantastically}, and some methods automate the generation of new examples \cite{margatina2023active} for improved performance.

\paragraph{Annotations}
Although bounding boxes, markers, and pixel masks are widely employed in vision-language tasks for visual guidance~\cite{kirillov2023segment}, these prompts remain exceptionally underexplored with respect to prompt optimization~\cite{gu2023systematic}. 
Existing approaches often treat the encoded image representation (e.g., image embeddings) as the ``soft'' prompt, leveraging soft prompt tuning strategies without explicitly capitalizing on human-crafted annotations. 
Only a limited body of work has introduced annotations as extra visual cues in manual prompt engineering~\cite{peng2023kosmos,denner2024visual}.

Though instructions, thoughts, exemplars, and annotations each serve distinct roles, many works integrate their optimization in a shared framework, aiming to maximize discrete prompt effectiveness while minimizing manual effort.

\subsection{Continuous Variables}\label{sec:soft-instruction}

In contrast to discrete tokens, continuous (soft) prompts ($\mathcal{P}_c$) rely on learnable embeddings ${\theta_i} \subset \mathbb{R}^d$. 
These vectors can be appended to input representations and optimized via gradient-based methods. 
By avoiding changes to the underlying model parameters, soft prompts require fewer resources for adaptation. 
Notable research explores \textit{prefix-based} \cite{li2021prefix,lester2021power,peng2025soft,wei2023improving} or \textit{layer-spanning embeddings} \cite{liu2024gpt}, revealing that tuning a small set of trainable vectors can achieve strong performance gains while preserving model generality. 
Soft prompts thus offer an efficient path to personalize or specialize foundation models without extensive fine-tuning or large-scale data.

\subsection{Mixed Variables}\label{sec:hybrid}

Mixed (hybrid) settings in $\mathcal{P}_h$ space incorporate both discrete and continuous elements, combining human-readable instructions or exemplars with trainable embedding vectors. 
This fusion leverages the interpretability and domain specificity of discrete tokens and the flexibility of continuous embeddings. 
Early works on co-optimizing instructions and exemplars \cite{fernando2023promptbreeder,wangmixture,agarwal2024promptwizard,cui2024phaseevo} demonstrate that jointly refining these complementary parts often achieves more robust and adaptable behavior than optimizing each component alone. 
Such synergy can be extended to multimodal prompts\cite{wen2024hard}, for instance, by coupling discrete spatial regions (in VLMs) with continuous alignment vectors, thereby enabling comprehensive, end-to-end optimization.

\section{Objective Functions}\label{sec: obj}

As formulated in Section~\ref{sec:formulation}, our goal is to solve
\begin{equation}\label{eq:obj-restate}
\max_{P \in \mathcal{P}} \;\; \mathbb{E}_{(x,y)\sim\mathcal{D}_{\text{val}}}\bigl[g\bigl(f\bigl(P(x)\bigr),y\bigr)\bigr],
\end{equation}
where $f\!:\!\mathcal{P}\times\mathcal{X}\to \mathcal{Y}$ denotes the black-box foundation model (either an LLM or VLM), $P$ is a prompt from the prompt space $\mathcal{P}$ (discrete, continuous, or hybrid), and $g:\mathcal{Y}\times\mathcal{Y}\to\mathbb{R}$ is a performance metric. 
We now illustrate how $g(\cdot)$ is instantiated across downstream tasks (Section~\ref{sec:downstream}) and then discuss constrained objectives (Section~\ref{sec:constrained}), all within the framework of Equation~\eqref{eq:main-formula}.

\subsection{Downstream Tasks}\label{sec:downstream}



\begin{figure}[htb!]
    \centering
    \includegraphics[width=\linewidth]{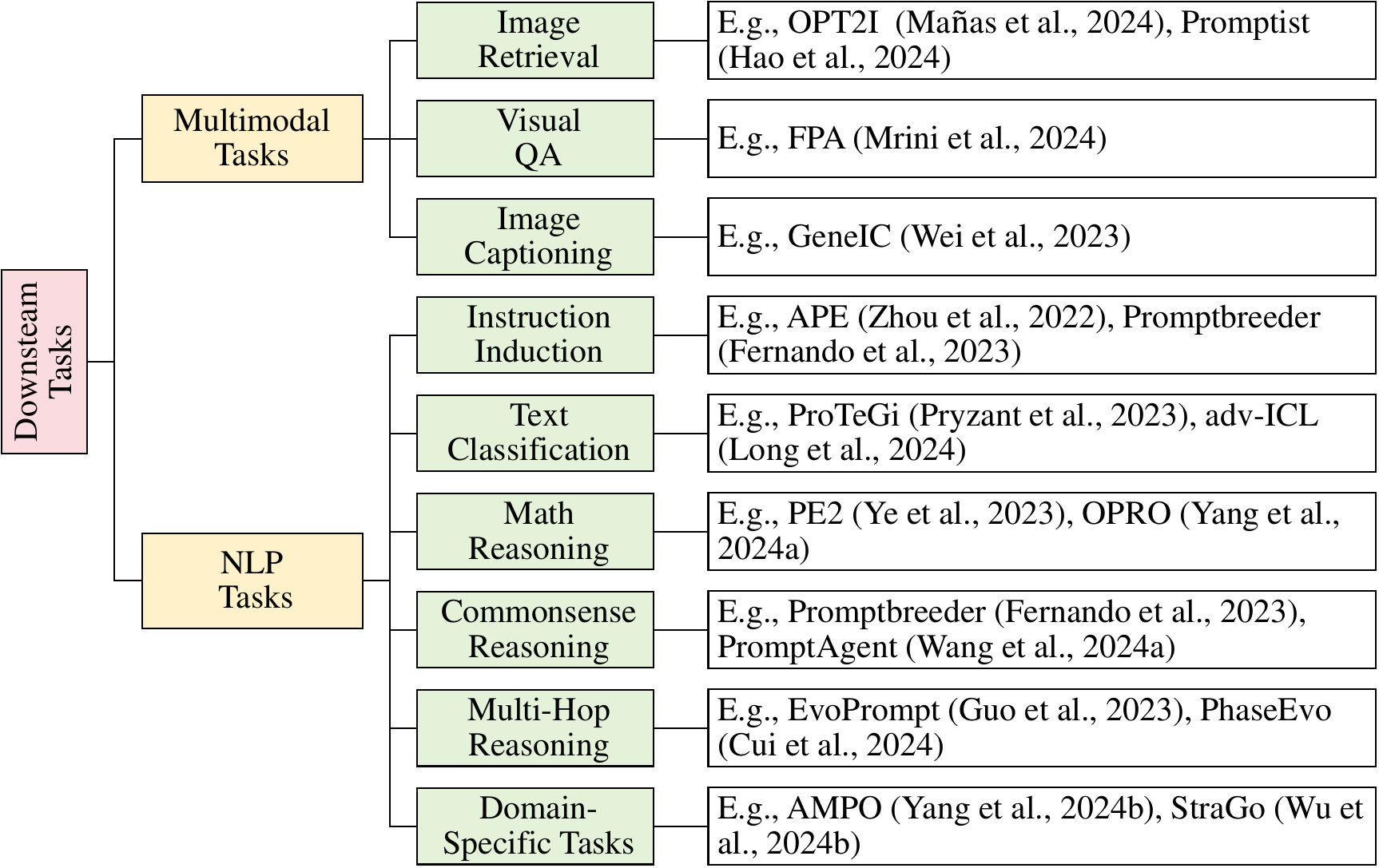}
    \caption{Taxonomy of Downstream Tasks and Corresponding Prompt Optimization Techniques.}
    \label{fig:downstream}
\end{figure}

\paragraph{Instruction Induction.}
This task evaluates how well $f\bigl(P(x)\bigr)$ extracts and generalizes underlying instructions. 
Following \cite{Honovich2022InstructionIF}, it spans 24 sub-tasks such as morphosyntactic transformations and causality detection. 
Typical $g(\cdot)$ measures include BERTScore-F1 and exact/set match.

\paragraph{Text Classification.}
Models map text $x\in\mathcal{X}$ to a discrete label $y\in\mathcal{Y}$. 
Datasets like  SST-2~\cite{socher2013recursive},  SST-5~\cite{socher2013recursive} and AGNEWS~\cite{zhang2015character} cover sentiment, topic classification, and subjectivity detection. 
Common metrics $g\bigl(f(P(x)),y\bigr)$ include classification accuracy and Macro-F1.

\paragraph{Math Reasoning.}
Here $f\bigl(P(x)\bigr)$ must solve arithmetic or algebraic word problems. 
Datasets such as GSM8K \cite{cobbe2021training}, MultiArith \cite{koncel2016mawps} and SingleEq \cite{koncel2016mawps} vary from single-step arithmetic to multi-step reasoning. 
Evaluation relies on comparing numeric outputs or perplexity.

\paragraph{Commonsense Reasoning.}
Models must incorporate external, everyday knowledge to answer or infer implicit context. Tasks like  CommonsenseQA~\cite{talmor2018commonsenseqa}, and StrategyQA~\cite{geva2021did} evaluate $f\bigl(P(x)\bigr)$ for multi-step common-sense inferences. 
Exact match are typical performance metrics.

\paragraph{Multi-Hop Reasoning.}
Extending beyond single-step inferences, multi-hop tasks demand chaining evidence from multiple sources. 
BBH~\cite{suzgun2022challenging} exemplifies this domain. 
The key objective is again $\max \mathbb{E}[g(\cdot)]$ over correctness but with specialized intermediate-step or CoT evaluations.

\paragraph{Domain-Specific Tasks.}
Focus on specialized areas demanding expert-level knowledge, e.g., biomedical tasks in MedQA~\cite{jin2021disease} and MedMCQA~\cite{pal2022medmcqa}, where $f\bigl(P(x)\bigr)$ must accurately handle topic-specific queries. 
Evaluation often uses accuracy or Macro-F1 on domain-specific labels.

\paragraph{Multimodal Tasks.}
Here $\mathcal{X}=\mathcal{X}_v\times\mathcal{X}_t$ includes image and text. 
Datasets MS~COCO~\cite{lin2014microsoft}, LAION~\cite{schuhmann2022laion}, and Celeb-A~\cite{liu2015deep} support image captioning, retrieval, and visual question answering. 
Objectives involve image-text alignment, retrieval accuracy, CLIP or Aesthetics Score.

\subsection{Constrained Objectives}\label{sec:constrained}

Beyond purely maximizing task-specific metrics, several prompt optimization scenarios require additional constraints:
\begin{itemize}
    \item \textbf{Prompt Editing.} Here we impose structural or semantic constraints on $P(x)\in \mathcal{P}$. Minor reformulations of instructions, exemplars, or spatial annotations must preserve or improve $\mathbb{E}[g(\cdot)]$ under restricted editing budgets. This suits tasks where $f(P(x))$ is sensitive to subtle prompt shifts \cite{sahoo2024systematic,amatriain2024prompt}.
    \item \textbf{Prompt Compression.} Constrains $\lVert P\!\rVert_{\text{length}} \leq \kappa$ (token-length or embedding-size budget), seeking $\max_{P\in\mathcal{P}} \mathbb{E}[g(\cdot)]$ subject to $\kappa$. Trimming extraneous tokens or embedding vectors increases computational efficiency while preserving accuracy, crucial in real-time or large-scale scenarios \cite{chang2024efficient}.
\end{itemize}

These can be viewed as special cases of Eq.~\eqref{eq:main-formula} with additional constraints $\Gamma(P)\leq \kappa$, leading to solutions that favor prompt conciseness or regulated edits. By integrating such constraints into the optimization procedure, automated prompt design expands beyond raw performance maximization to meet usability, efficiency, and alignment requirements.

\section{Optimization Methods}\label{sec: method}

\begin{figure}[htb!]
    \centering
    \includegraphics[width=.6\linewidth]{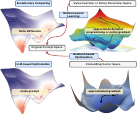}
    \caption{The landscapes of different optimization methods.}
    \label{fig:landscape}
\end{figure}

To establish a cohesive organization, we propose a unifying taxonomy and classify optimization methods across four major paradigms, as shown in Figure~\ref{fig:landscape}: 
(1) FM-based Optimization, (2) Evolutionary Computing, (3) Gradient-Based Optimization, and (4) Reinforcement Learning. 
Each paradigm can further be subdivided based on whether it optimizes purely discrete prompts ($P\in\mathcal{P}_d$), purely continuous prompts ($P\in\mathcal{P}_c$), or hybrids ($P\in\mathcal{P}_h$). 
Many works mix or extend these categories (e.g., \emph{EvoPrompt} combines FM-based generation with Genetic Algorithm (GA) operators). 
By situating each study within this two-dimensional taxonomy, we highlight shared structural principles across seemingly diverse algorithms and providing a cohesive view of the rapidly expanding literature, as shown in Table~\ref{tab:category}.

\subsection{FM-based Optimization}

FM-based optimization methods directly leverage FM as \emph{meta-optimizers} to refine prompts. 
These approaches often implement iterative improvement in which an FM proposes an updated prompt based on performance feedback.

\paragraph{Heuristic Meta-Prompt}
Several methods harnesses human-designed meta-prompts, i.e., manually-craft sequences that instruct an FM how to revise an existing prompt.
\emph{PE2}~\cite{ye2023prompt} uses rich meta-descriptions, context specifications, and CoT templates to iteratively update prompts for various tasks. 
\emph{OPRO}~\cite{yang2023largelm} unifies solution exploration and evaluation, integrating previously generated solutions (and their quality metrics) within a meta-prompt that the FM uses to refine future versions of $P$. 
\emph{LCP}~\cite{li2024learning} integrates contrastive learning signals into meta-prompts, encouraging FMs to distinguish high-quality $P_d\in\mathcal{P}_d$ from suboptimal ones, and adapt across model families/languages. 
Similarly, \emph{StraGo}~\cite{wu2024strago} merges success and failure exemplars as meta-context to steer in-context learning toward more robust prompts.

\paragraph{Automatic Meta-Prompt Generation}
Another subset generate meta-prompt relying on external feedback and self-reflection. 
\emph{ProTeGi}~\cite{pryzant2023automatic} formulates an iterative ``gradient-like'' textual editing loop, incorporating beam search and multi-armed bandit strategies to refine discrete prompts $P\in\mathcal{P}_d$. 
\emph{AutoHint}~\cite{sun2023autohint} appends FM-inferred hints derived from prior prediction errors, thereby evolving initial prompts in a step-by-step manner. 
\emph{CriSPO}~\cite{he2024crispo} introduces critique-suggestion pairs that guide FM feedback to improve text generation prompts without modifying model weights; it further proposes Automatic Suffix Tuning (AST) for multi-objective prompt engineering. 
Likewise, \emph{BPO}~\cite{cheng2023black} aligns $f(P(x))$ with user intent by collecting human feedback on interim outputs, then editing $P$ accordingly.

\paragraph{Strategic Search and Replanning}
A few works incorporate explicit search strategies. 
\emph{APE}~\cite{zhou2022large} conducts black-box prompt exploration through an FM-proposed candidate pool, selecting prompts that maximize task performance without requiring gradients. 
\emph{PromptAgent}~\cite{wang2023promptagent} leverages Monte Carlo Tree Search (MCTS) to navigate a combinatorial space of expert-level prompts, applying user feedback as value signals. 
\emph{AMPO}~\cite{yang2024ampo} evolves multi-branched prompts, guided by failures and partial successes; each branch refines $P(x)$ to better handle increasingly complex task variations. 
\emph{adv-ICL}~\cite{long2024prompt} employs a generator-discriminator FM setup to explore adversarial prompts, leading to robust in-context demonstrations. 
\emph{OPT2I}~\cite{manas2024improving} focuses on text-to-image consistency by rewriting textual prompts $P_d$ to boost alignment.

\begin{table}[ht]
\centering
\resizebox{\textwidth}{!}{%
\begin{tabular}{lllll}
\hline
\textbf{Method} & \textbf{Optimization Space} & \textbf{Variable Type} & \textbf{Optimization Methods} & \textbf{Optimization Strategy} \\ \hline
PE2~\cite{ye2023prompt} & Discrete & Instructions & FM-based Optimization & Heuristic Meta-Prompt \\ 
OPRO~\cite{yang2023largelm} & Discrete & Instructions & FM-based Optimization & Heuristic Meta-Prompt \\ 
LCP~\cite{li2024learning} & Discrete & Instructions & FM-based Optimization & Heuristic Meta-Prompt \\ 
StraGo~\cite{wu2024strago} & Discrete & Instructions & FM-based Optimization & Heuristic Meta-Prompt \\ 
ProTeGi~\cite{pryzant2023automatic} & Discrete & Instructions &  FM-based Optimization & Automatic Meta-Prompt Generation \\ 
AutoHint~\cite{sun2023autohint} & Discrete & Instructions &  FM-based Optimization & Automatic Meta-Prompt Generation \\ 
CriSPO~\cite{he2024crispo} & Discrete & Instructions &  FM-based Optimization & Automatic Meta-Prompt Generation \\ 
BPO~\cite{cheng2023black} & Discrete & Instructions & FM-based Optimization & Automatic Meta-Prompt Generation \\ 
APE~\cite{zhou2022large} & Discrete & Instructions &  FM-based Optimization & Strategic Search and Replanning \\ 
PromptAgent~\cite{wang2023promptagent} & Discrete & Instructions &  FM-based Optimization & Strategic Search and Replanning \\ 
AMPO~\cite{yang2024ampo} & Discrete & Instructions & FM-based Optimization & Strategic Search and Replanning \\ 
adv-ICL~\cite{long2024prompt} & Discrete & Instructions & FM-based Optimization & Strategic Search and Replanning \\ 
OPT2I~\cite{manas2024improving} & Discrete & Instructions & FM-based Optimization & Strategic Search and Replanning \\ 
GPS~\cite{xu2022gps} & Discrete & Instructions & Evolutionary Computing & Genetic Operators and Heuristics \\ 
LongPO~\cite{hsieh2023automatic} & Discrete & Instructions & Evolutionary Computing & Genetic Operators and Heuristics \\ 
GrIPS~\cite{prasad2022grips} & Discrete & Instructions & Evolutionary Computing & Genetic Operators and Heuristics \\ 
EvoPrompt~\cite{guo2023connecting} & Discrete & Instructions & Evolutionary Computing & Self-Referential Evolution \\ 
HPME~\cite{wen2024hard} & Discrete & Instructions & Gradient-Based Optimization & Discrete Token Gradient Methods \\ 
AutoPrompt~\cite{shin2020autoprompt} & Discrete & Instructions & Gradient-Based Optimization & Discrete Token Gradient Methods \\ 
ZOPO~\cite{hu2024localized} & Discrete & Instructions & Gradient-Based Optimization & Discrete Token Gradient Methods \\ 
RLPrompt~\cite{deng2022rlprompt} & Discrete & Instructions & Reinforcement Learning & Prompt Editing as RL Actions \\ 
TEMPERA~\cite{zhang2022tempera} & Discrete & Instructions & Reinforcement Learning & Prompt Editing as RL Actions \\ 
PRewrite~\cite{kong2024prewrite} & Discrete & Instructions & Reinforcement Learning & Prompt Editing as RL Actions \\ 
PACE~\cite{dong2023pace} & Discrete & Instructions & Reinforcement Learning & Prompt Editing as RL Actions \\ 
StablePrompt~\cite{kwon2024stableprompt} & Discrete & Instructions & Reinforcement Learning & Prompt Editing as RL Actions \\ 
Evoke~\cite{hu2023evoke} & Discrete & Instructions & Reinforcement Learning & Prompt Editing as RL Actions \\ 
Prompt-OIRL~\cite{sun2023query} & Discrete & Instructions & Reinforcement Learning & Multi-Objective and Inverse RL Strategies \\ 
MORL-Prompt~\cite{jafari2024morl} & Discrete & Instructions & Reinforcement Learning & Multi-Objective and Inverse RL Strategies \\ 
MAPO~\cite{chen2024mapo} & Discrete & Instructions & Reinforcement Learning & Multi-Objective and Inverse RL Strategies \\
Self-ask~\cite{press2023measuring} & Discrete & Thoughts & FM-based Optimization & Strategic Search and Replanning \\
Reprompt~\cite{chen2024reprompt} & Discrete & Thoughts  & FM-based Optimization & Strategic Search and Replanning \\
PROMST~\cite{chen2024prompt} & Discrete & Thoughts & Evolutionary Computing & Self-Referential Evolution \\ 
ReAcT~\cite{yaoreact} & Discrete & Thoughts & FM-based Optimization & Strategic Search and Replanning \\
Self-refine~\cite{madaan2024self} & Discrete & Thoughts &  FM-based Optimization & Strategic Search and Replanning \\ 
KATE~\cite{liu2022makes} & Discrete & Few-shot Examples &  FM-based Optimization & Strategic Search and Replanning \\ 
AL~\cite{margatina2023active} & Discrete & Few-shot Examples & FM-based Optimization & Strategic Search and Replanning \\ 
AES~\cite{zhang2022active} & Discrete & Few-shot Examples & Reinforcement Learning & Prompt Editing as RL Actions \\ 
Ordered Prompt~\cite{lu2022fantastically} & Discrete & Few-shot Examples & FM-based Optimization & Strategic Search and Replanning \\
SA~\cite{kirillov2023segment} & Discrete & Annotations &  FM-based Optimization & Automatic Meta-Prompt Generation\\
KOSMOS-2~\cite{peng2023kosmos} & Discrete & Annotations & FM-based Optimization & Automatic Meta-Prompt Generation\\
Visualcues~\cite{denner2024visual} & Discrete & Annotations & FM-based Optimization & Automatic Meta-Prompt Generation\\
Prefix-tuning~\cite{li2021prefix} & Continuous & Soft Prompt & Gradient-Based Optimization & Soft Prompt Tuning \\ 
Prompt-Tuning~\cite{lester2021power} & Continuous & Soft Prompt & Gradient-Based Optimization & Soft Prompt Tuning \\ 
P-Tuning~\cite{liu2024gpt} & Continuous & Soft Prompt & Gradient-Based Optimization & Soft Prompt Tuning \\ 
PhaseEvo~\cite{cui2024phaseevo} & Hybrid & Instructions and Exemplars & Evolutionary Computing & Genetic Operators and Heuristics \\ 
Promptbreeder~\cite{fernando2023promptbreeder} & Hybrid & Instructions and Exemplars & Evolutionary Computing & Self-Referential Evolution \\ 
Mixture-of-Prompts~\cite{wangmixture} & Hybrid & Instructions and Exemplars &  FM-based Optimization & Automatic Meta-Prompt Generation \\
Promptwizard~\cite{agarwal2024promptwizard} & Hybrid & Instructions and Exemplars & FM-based Optimization & Automatic Meta-Prompt Generation \\ 
\hline

\end{tabular}
}
\caption{Categorization of representive papers based on optimization variables and methods.}
\label{tab:category}
\end{table}

\subsection{Evolutionary Computing}


Evolutionary methods model prompt optimization as a genetic or evolutionary process. 
They treat the prompt $P\in\mathcal{P}_d$ as an ``organism,'' mutated or crossed over to produce ``offspring'' that survive based on higher fitness (i.e., the performance measure $g(f(P(x)),\,y)$). 
These approaches are particularly suitable for purely discrete prompt spaces.

\paragraph{Genetic Operators and Heuristics}
\emph{GPS}~\cite{xu2022gps} applies a straightforward genetic algorithm to refine few-shot instruction prompts, iteratively mutating tokens and retaining top-performing discrete prompts. 
\emph{LongPO}~\cite{hsieh2023automatic} extends these ideas to long prompts by incorporating beam search heuristics and a history buffer to preserve context across mutations. 
Though not strictly a GA, \emph{GrIPS}~\cite{prasad2022grips} shares a similar local-edit mechanism to produce improved child prompts from parent instructions. 
\emph{PhaseEvo}~\cite{cui2024phaseevo} further unifies instruction and example optimization into a multi-phase generation pipeline, alternating between refining $I,\,T\in\mathcal{V}^*$ and exemplar sets $\{e_i\}_{i=1}^k\subseteq\mathcal{E}$.

\paragraph{Self-Referential Evolution}
Other approaches tap into FMs themselves as evolutionary operators. 
\emph{EvoPrompt}~\cite{guo2023connecting} uses the FM to propose candidate mutations, combining them with a fitness-based selection procedure. 
\emph{Promptbreeder}~\cite{fernando2023promptbreeder} co-evolves \emph{both} task prompts and the mutation prompts: the latter are instructions specifying how to mutate or cross over parent prompts. 
\emph{PROMST}~\cite{chen2024prompt} specializes in multi-step tasks. 
It incorporates a human-in-the-loop rule set and a learned heuristic model, mixing evolutionary sampling with user feedback to gradually improve the textual prompt.

\subsection{Gradient-Based Optimization}

Gradient-based strategies derive from classical optimization principles, but face unique obstacles in prompt engineering since discrete tokens $\bigl[I,T,\{e_i\}\bigr]\in\mathcal{P}_d$ are not directly differentiable. 
Methods in this family either approximate gradients to navigate discrete spaces or, more commonly, optimize continuous parameters $\theta\in\mathbb{R}^d$ in a soft prompt context $P\in\mathcal{P}_c$.

\paragraph{Discrete Token Gradient Methods}
For closed-source FMs, direct gradient access is often unavailable, requiring alternative solutions:
\emph{HPME}~\cite{wen2024hard} projects a learned continuous embedding back to discrete tokens each iteration, blending soft gradient updates with nearest-neighbor token matching. 
\emph{AutoPrompt}~\cite{shin2020autoprompt} constructs prompts by adding tokens that maximize the gradient toward correct labels in a masked language modeling scenario. 
\emph{ZOPO}~\cite{hu2024localized} implements zeroth-order optimization in discrete prompt spaces by sampling localized perturbations in the token domain, guided by a neural tangent kernel approximation.

\paragraph{Soft Prompt Tuning}
Soft prompt methods treat $P\in\mathcal{P}_c$ as a set of trainable vectors $\{\theta_1,\dots,\theta_m\}$ that concatenate with the embedding of $x$, i.e., $P(e_x)=[\theta_1,\ldots,\theta_m; e_x]$. 
\emph{Prefix-tuning}~\cite{li2021prefix} attaches learnable prefix vectors in the hidden states of a language model, requiring only a small fraction of trainable parameters. 
\emph{Prompt-Tuning}~\cite{lester2021power} similarly adds trainable embeddings at the input layer, benefiting from large model scalability. 
\emph{P-Tuning}~\cite{liu2024gpt} extends trainable prompts into multiple layers, significantly improving few-shot performance. 
All these methods solve variants of the continuous optimization subproblem $\max_{P\in\mathcal{P}_c}\;\mathbb{E}_{(x,y)\sim\mathcal{D}_{\text{val}}}[g(f(P(e_x)),\,y)]$
and hence leverage standard gradient descent w.r.t. $\theta_i\in\mathbb{R}^d$.

\subsection{Reinforcement Learning}

RL methods recast prompt design as an RL problem in which $P\in\mathcal{P}$ (often $\mathcal{P}_d$ or $\mathcal{P}_h$) is updated via a sequence of actions under a reward defined by $g\bigl(f(P(x)),\,y\bigr)$.
Across these RL methods, the key idea is to formulate optimization objective $
\max_{P\in\mathcal{P}}\;\mathbb{E}_{(x,y)\sim\mathcal{D}_{\text{val}}}[g(f(P(x)),y)]$
as a Markov Decision Process, where partial or recurrent prompt edits constitute actions and $g$ serves either directly as the reward function or as part of a learned proxy. 
Such frameworks unify discrete prompt editing ($P\in\mathcal{P}_d$) and continuous prompt adaptation ($P\in\mathcal{P}_c$), even allowing for hybrid forms $P\in\mathcal{P}_h$ where some components are differentiable (\emph{soft prompts}) and others remain discrete \emph{(hard instructions)}.

\paragraph{Prompt Editing as RL Actions}
\emph{RLPrompt}~\cite{deng2022rlprompt} represents discrete tokens $v\in\mathcal{V}$ as RL actions, exploring the space of textual prompts with policy gradient methods. 
\emph{TEMPERA}~\cite{zhang2022tempera} proposes test-time RL-based editing, adjusting each query's prompt adaptively. 
\emph{PRewrite}~\cite{kong2024prewrite} trains a separate prompt rewritter using RL signals to maximize downstream task accuracy. 
In a similar vein, \emph{PACE}~\cite{dong2023pace} refines suboptimal human prompts ($P_{\text{human}}$) by iterative RL feedback, while \emph{StablePrompt}~\cite{kwon2024stableprompt} adapts proximal policy optimization to mitigate training instability. 
\emph{Evoke}~\cite{hu2023evoke} establishes a reviewer-author loop, feeding prompt outputs into a critic that suggests incremental edits.

\paragraph{Multi-Objective and Inverse RL Strategies}
Other RL approaches tackle multi-objective or partial feedback scenarios. 
\emph{Prompt-OIRL}~\cite{sun2023query} employs offline inverse RL to learn a query-specific reward model, thus selecting an optimal prompt without frequent FM interactions. 
\emph{MORL-Prompt}~\cite{jafari2024morl} addresses conflicting reward functions (e.g., style vs.\ accuracy) by adapting multi-objective RL techniques. 
\emph{MAPO}~\cite{chen2024mapo} combines supervised fine-tuning and RL in a model-adaptive prompt optimizer that tailors $P$ to each target FM, demonstrating notable gains across diverse downstream tasks.

\section{Future Directions}\label{sec:future}

The systematic study of prompt optimization for foundation models, from an applied optimization perspective, presents extensive opportunities but remains loosely explored. 
Below, we outline key research themes, highlighting current progress, pivotal questions, and open challenges.

\paragraph{Constraint optimization}
Presents methods seldom incorporate semantic or ethical constraints in discrete prompt spaces. 
The main issue lies in constructing a search mechanism that respects human-value alignment, resource bounds, and readability, particularly in high-dimensional symbolic domains. 
Another challenge is formalizing these constraints as tractable mathematical conditions that guide search algorithms without sacrificing flexibility or linguistic quality.

\paragraph{Multi-task prompt optimization}
It's crucial for leveraging shared structures across tasks. 
While some work suggests factorized or sparse representations to capture prompt-level similarity, formal definitions of inter-task ``prompt similarity'' are lacking. 
Negative transfer can also arise, where improvements for one task degrade performance on another. 
The field needs robust frameworks to codify these trade-offs, enhancing generalization and adaptivity.

\paragraph{Online prompt optimization}
Current techniques generally prioritize offline scenarios. 
However, user intentions can shift over time, creating the need for algorithms that maintain stable performance (e.g., bounded dynamic regret) in nonstationary environments. 
Reliance on online updates amplifies the complexity of discrete search in high-dimensional prompt spaces. 
Furthermore, real-time user feedback loops introduce additional uncertainties, demanding advanced convergence analyses.

\paragraph{Multi-objective prompt optimization} 
Multi-objective prompt optimization aims to balance often competing goals such as accuracy and interpretability. 
Many existing studies use single-metric optimization, overlooking fundamental human-centered preferences. 
One promising direction is the incorporation of Pareto-based methods or multi-criteria decision-making, accompanied by geometric representations of preference spaces. 
Game-theoretic techniques may also help arbitrating among conflicting objectives, both within and across user populations.

\paragraph{Heterogeneous modality optimization}
Although many works focus on textual cues, prompts in computer vision and other modalities, such as bounding boxes or pixel-level annotations, remain far less explored. 
This calls for a deeper understanding of cross-modal coupling, as well as conditions under which modalities are amenable to joint or separate optimization. 
Such advances might require novel manifold-based or graph-based tools to unify distinct prompt representations.

\paragraph{Bi-level prompt optimization} 
Bi-level prompt optimization arises with step-by-step ``thought-driven'' models (e.g., OpenAI-o1, Deepseek-R1), where the entire inference chain depends on the prompt as a high-level controller. 
This hierarchical structure challenges standard prompt optimization, as small changes can drastically alter the reasoning trajectory. 
It remains open whether stable equilibria for these layered systems exist and how sensitive they are to prompt perturbations. 
Borrowing methods from multi-level optimization could clarify both existence and uniqueness conditions for equilibria.

\paragraph{Broader application scenarios}
For multi-turn agents, prompt optimization unfolds over sequential decisions, significantly complicating nonstationarity. 
Introducing game-theoretic elements for multi-agent designs further underscores the importance of collaborative or competitive equilibrium concepts. 
Similarly, vertical-domain large models (e.g., reinforcement learning, AI4Science) impose domain-specific constraints that standard prompt optimization does not account for, calling for specialized theoretical adaptations.

\section{Discussion and Conclusion}\label{sec:conclusion}


This survey delineates a optimization-theoretic foundation for automated prompt engineering that transcends fragmented treatments across modalities.  
By synthesizing methods that target discrete, continuous, and hybrid prompt spaces, we have underscored how variables such as instructions, soft prompts, and exemplars can be systematically optimized under unified theoretical principles. 
Together with our taxonomy of task objectives and unified perspective on FM as optimizer, evolutionary computing, gradient-based, and RL-driven methods, we establish key foundations for theoretical inquiry and realistic application. 
Moving forward, tighter integration of multi-level, multi-objective, and online optimization will be pivotal in shaping prompt designs for emerging foundation models.

\clearpage
\newpage

\bibliographystyle{alpha}
\bibliography{main2}
\end{document}